\documentclass{easychair}

\usepackage{graphicx}
\usepackage{booktabs}
\usepackage{amssymb}
\usepackage{makecell}
\usepackage{multirow}
\usepackage{subcaption}
\usepackage{amsmath}
\usepackage{mathtools}
\usepackage{tabularx}
\usepackage{algpseudocode}
\usepackage{algorithm}
\usepackage[subtle]{savetrees}
\usepackage[T1]{fontenc}
\usepackage{lmodern}
\usepackage{xcolor}
\usepackage{doc}

\title{ANTONIO: Towards a Systematic Method for Generating NLP Benchmarks for Verification}

\author{
	Marco Casadio\inst{1}\and
	Luca Arnaboldi\inst{2}\thanks{Large portion of work undertaken whilst at University of Edinburgh}\and
	Matthew L. Daggitt\inst{1}\and
	Omri Isac\inst{3}\and
	Tanvi Dinkar\inst{1}\and
	Daniel Kienitz\inst{1}\and
	Verena Rieser \inst{1}\and
	Ekaterina Komendantskaya\inst{1}
	}
\institute{
	Heriot-Watt University, Edinburgh, UK\\
	\email{\{mc248,md2006,t.dinkar,dk50,v.t.rieser,e.komendantskaya\}@hw.ac.uk}\and
	University of Birmingham, Birmingham, UK\\
	\email{l.arnaboldi@bham.ac.uk}\and
	The Hebrew University of Jerusalem, Jerusalem, Israel\\
	\email{omri.isac@mail.huji.ac.il}
	}

\authorrunning{Casadio, M et al.}

\titlerunning{ANTONIO: Generating NLP Benchmarks for Verification}

\begin{document}
\maketitle

\begin{abstract}

Verification of machine learning models used in Natural Language Processing (NLP) is known to be a hard problem. In particular, many known neural network verification methods that work for computer vision and other numeric datasets do not work for NLP. Here, we study technical reasons that underlie this problem. Based on this analysis, we propose practical methods and heuristics for preparing NLP datasets and models in a way that renders them amenable to state-of-the-art verification methods. We implement these methods as a Python library called ANTONIO that links to the neural network verifiers ERAN and Marabou. We perform evaluation of the tool using an NLP dataset R-U-A-Robot suggested as a benchmark for verifying legally critical NLP applications. We hope that, thanks to its general applicability, this work will open novel possibilities for including NLP verification problems into neural network verification competitions, and will popularise NLP problems within this community.

\end{abstract}

\section{Introduction}
\label{sec:section1}


Deep neural networks (DNNs) are adept at addressing challenging problems in various areas, such as Computer Vision (CV)~\cite{ren2016faster} and Natural Language Processing (NLP)~\cite{sutskever2014sequence,advancesNLP}.
Due to their success, systems based on DNNs are widely deployed in the physical world and their safety and security is a critical matter~\cite{10.1145/3442188.3445922,RisksofFoundationModels,E2ECAI}.

One example of a safety critical application in the NLP domain is a chatbot's responsibility to identify itself as an AI agent, \emph{when asked by the user to do so}. Recently there has been several pieces of legislation proposed that will enshrine this requirement in law~\cite{EUlaw,CAlaw}.
For the chatbot to be compliant with these new laws, the DNN, or the subsystem responsible for identifying these queries, must be 100\% accurate in its recognition of the user's question. Yet, in reality the questions can come in different forms, for example: \emph{``Are you a Robot?''}, \emph{``Am I speaking with a person?''},  \emph{``Hey, are you a human or a bot?''}. Failure to recognise the user's intent and thus failure to answer the question correctly can have legal implications for the chatbot designers~\cite{EUlaw,CAlaw}.

The R-U-A-Robot dataset~\cite{gros2021ruarobot} was created with the intent to study solutions to this problem, and prevent user discomfort or deception in situations where users might have unsolicited conversations with human-sounding machines over the phone. It contains $6800$ sentences of this kind, labelled as positive (the question demands identification), negative (the question is about something else) and ambiguous. One can train and use a DNN to identify the intent. 

How difficult is it to formally guarantee the DNN's intended behaviour? The state-of-the-art NLP technology usually relies on \emph{transformers}~\cite{attentionisallyouneed} to embed natural language sentences into vector spaces.
Once this is achieved, an additional medium-size network may be trained on a dataset that contains different examples of \emph{``Are you a Robot?''} sentences.
This second network will be used to classify new sentences as to whether they contain the intent to ask whether the agent is a robot.

There are several approaches to verify such a system. Ideally we would try to verify the whole system consisting of both the embedding function and the classifier. However, state-of-the-art transformers are beyond the reach of state-of-the-art verifiers. 
For example the base model of BERT~\cite{bert} has around 110 million trainable parameters and GPT-3~\cite{gpt3} has around 175 billion trainable parameters.
In contrast, the most performant neural network verifier AlpaBetaCrown~\cite{wang2021beta} can only handle networks in the range of a few hundred of thousand nodes.
So, verification efforts will have to focus on the classifier that stands on top of the transformer.  

Training a DNN with 2 layers on the R-U-A-Robot dataset~\cite{gros2021ruarobot} gives average accuracy of 93\%.  Therefore there is seemingly no technical hurdle in running existing neural network verifiers on it~\cite{marabou,singh2019abstract,wang2021beta}. However, most of the properties checked by these verifiers are in the computer vision domain. In this domain, images are seen as vectors in a continuous space, and every point in the space corresponds to a valid image. The act of verification guarantees that every point in a given region of that space is classified correctly. Such regions are identified as ``$\epsilon$-balls'' drawn around images in the training dataset, for some given constant $\epsilon$. Despite not providing a formal guarantee about the entire space, this result is useful as it provides guarantees about the behaviour of the network over a large set of unseen inputs.

\begin{figure}[t]
\centering
	\begin{subfigure}[b]{0.3\textwidth}
		\includegraphics[width=\textwidth]{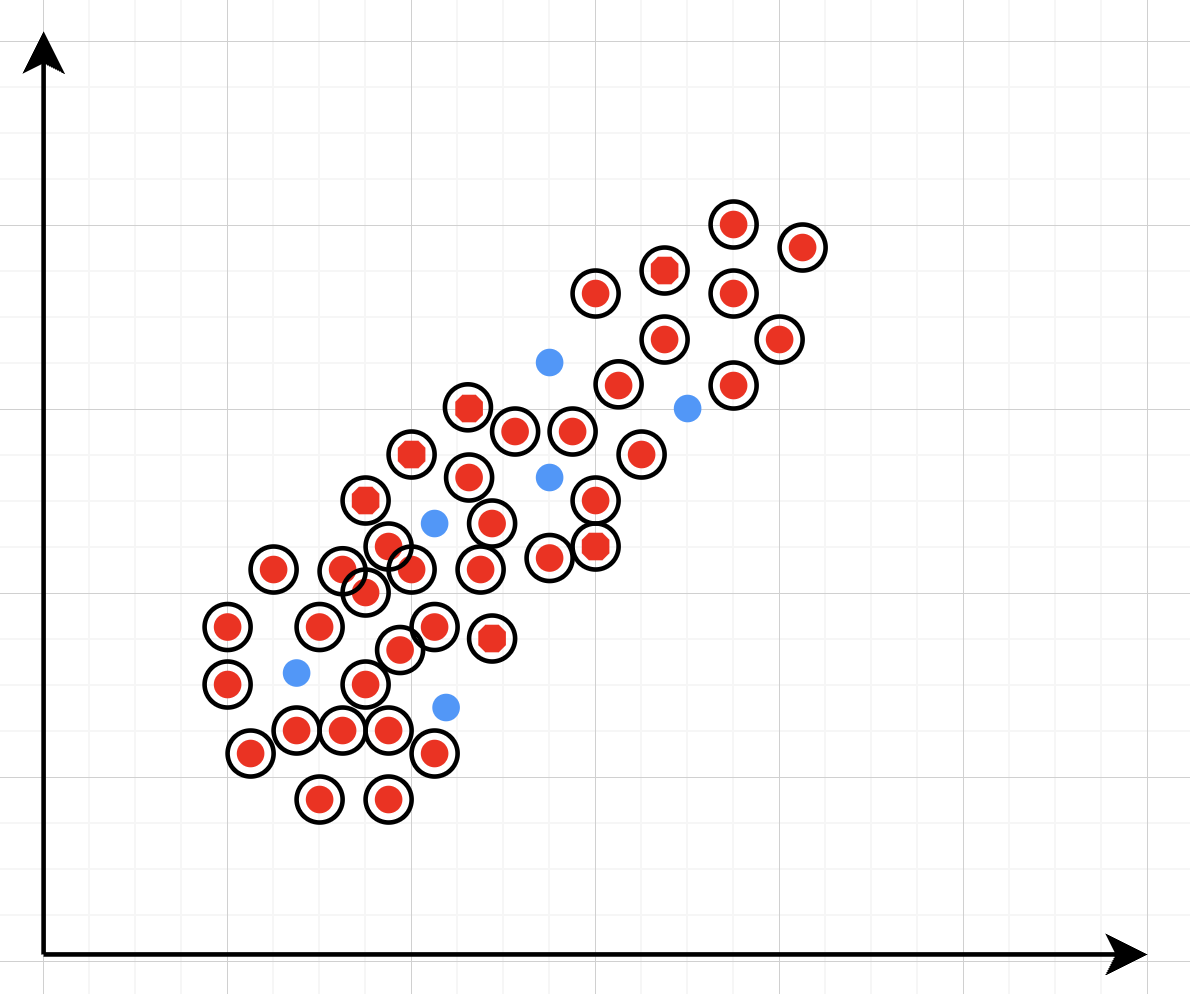}
	\end{subfigure}
	\begin{subfigure}[b]{0.3\textwidth}
		\includegraphics[width=\textwidth]{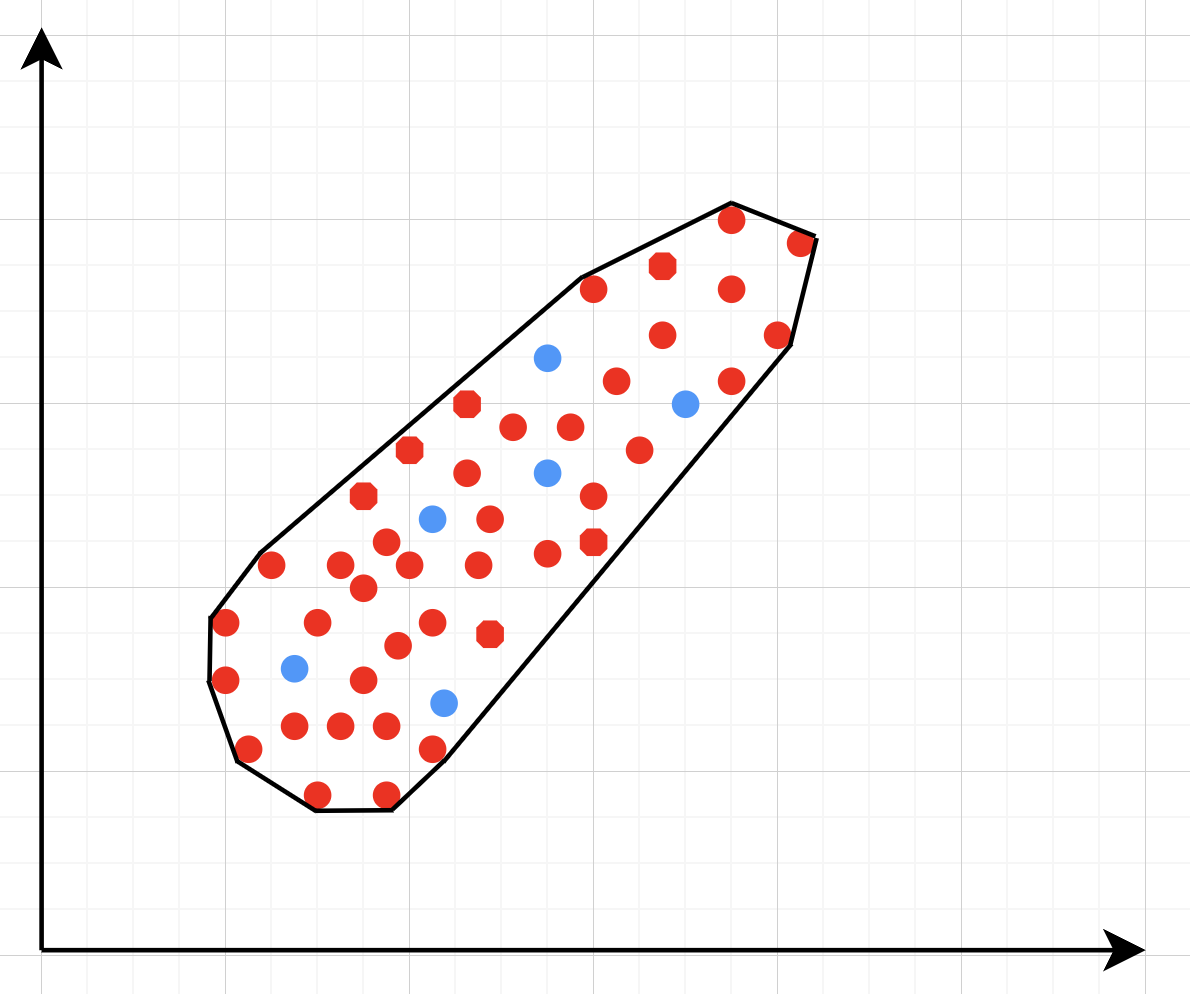}
	\end{subfigure}
	\begin{subfigure}[b]{0.3\textwidth}
		\includegraphics[width=\textwidth]{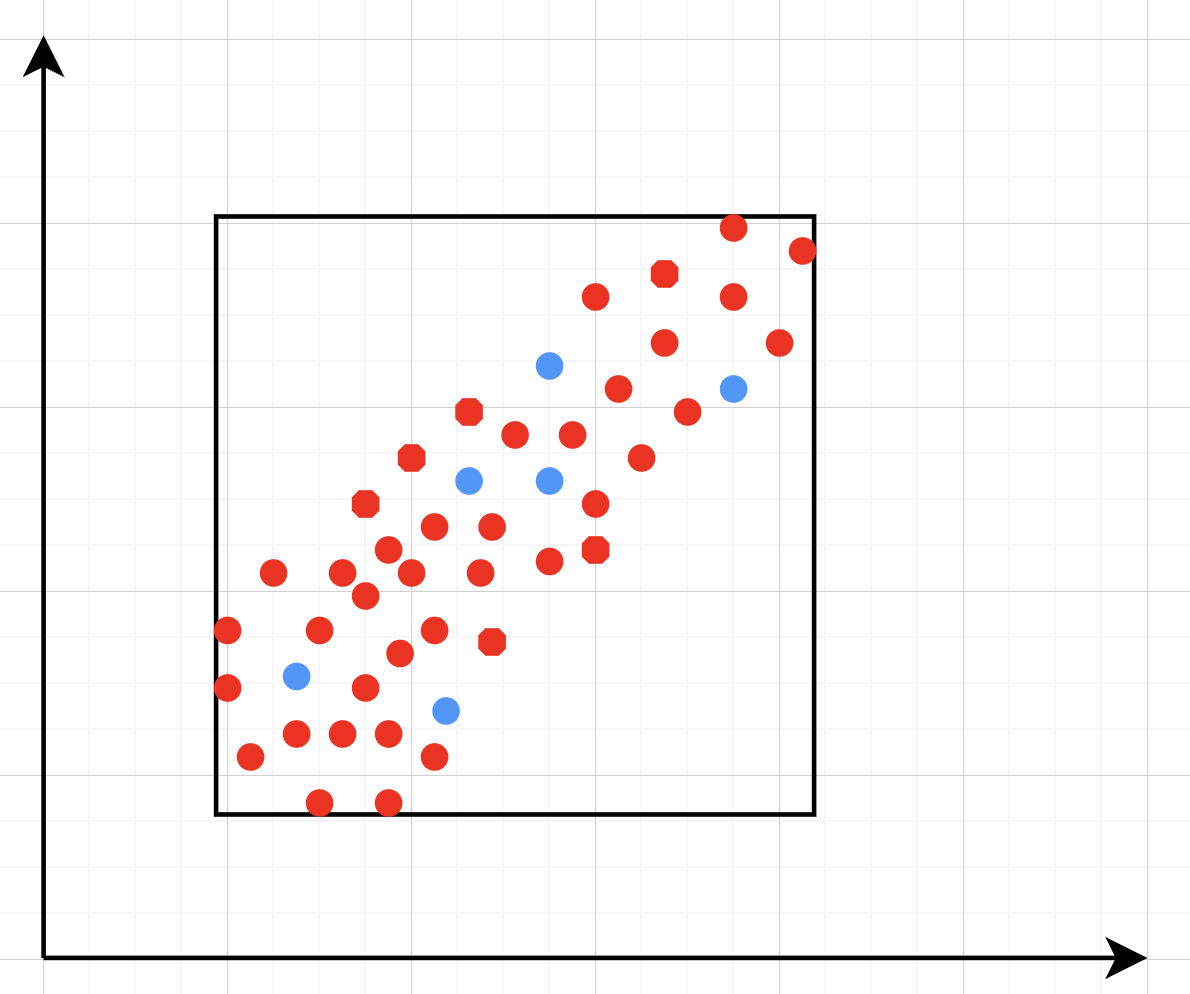}
	\end{subfigure}
\caption{\emph{An example of $\epsilon$-balls (left), convex-hull (centre) and hyper-rectangle (right) in 2-dimensions. The red dots represent sentences in the embedding space from the training set belonging to one class, while the torquoise dots are embedded sentences from the test set belonging to the same class.}}
\label{fig:ball-hrect}
\end{figure}

However, if we replicate this approach in the NLP domain, we obtain a mathematically sound but pragmatically useless result. This is because, unlike images, sentences form a discrete domain, and therefore very few points in the input space actually correspond to valid sentences. Therefore, as shown in Figure~\ref{fig:ball-hrect}, it is highly unlikely that the $\epsilon$-balls will contain any unseen sentences for values of $\epsilon$ that can actually be verified. And thus, such verification result does not give us more assurance than just measuring neural network accuracy!  

There is clearly a need in a substantially different methodology for verification of NLP models. Our proposal is based on the following observations.
On the verifier side, the state-of-art tools based on abstract interpretation are still well-suited for this task, because they are flexible in the definition of the size and shape of 
the input sub-space that they verify. But considerable effort needs to be put into definitions of subspaces that actually make pragmatic sense from the NLP perspective.
For example, as shown in Figure~\ref{fig:ball-hrect}, constructing a convex hull around several sentences embedded in the vector space has a good chance of capturing new, yet unseen sentences. 

Unfortunately, calculating convex hulls with sufficient precision is computationally infeasible for high number of dimensions. We resort to over-approximating convex hulls with   
\emph{``hyper-rectangles''}, computation of which only takes into consideration the minimum and maximum value of each dimension for each point around which we draw the hyper-rectangle. 
There is one last hurdle to overcome: just naively drawing hyper-rectangles around some data points gives negative results, i.e. verifiers fail to prove the correctness of classifications within the resulting hyper-rectangles.

There is no silver bullet to overcome this problem. Instead, as we show in this paper, one needs a systematic methodology to overcome this problem.
Firstly, we need methods that refine hyper-rectangles in a variety of ways, from the standard tricks of reducing dimensions of the embedding space and clustering, 
to geometric manipulations, such as hyper-rectangle rotation. 
Secondly, precision of the hyper-rectangle shapes can be improved by generating valid sentence perturbations, and constructing hyper-rectangles around (embeddings of) perturbed, and thus semantically similar, sentences. 
Finally, based on the refined spaces, we must be able to re-train the neural networks to correctly fit the shapes, 
and this may involve sampling based on adversarial attacks within the hyper-rectangles. This final step is in line with other literature in this domain; an approach that couples verification with property-driven training is also known as continuous verification~\cite{KomendantskayaK20,CKDKKAE22,SKDSS23}.

\begin{figure}[htbp]
\includegraphics[width=\textwidth]{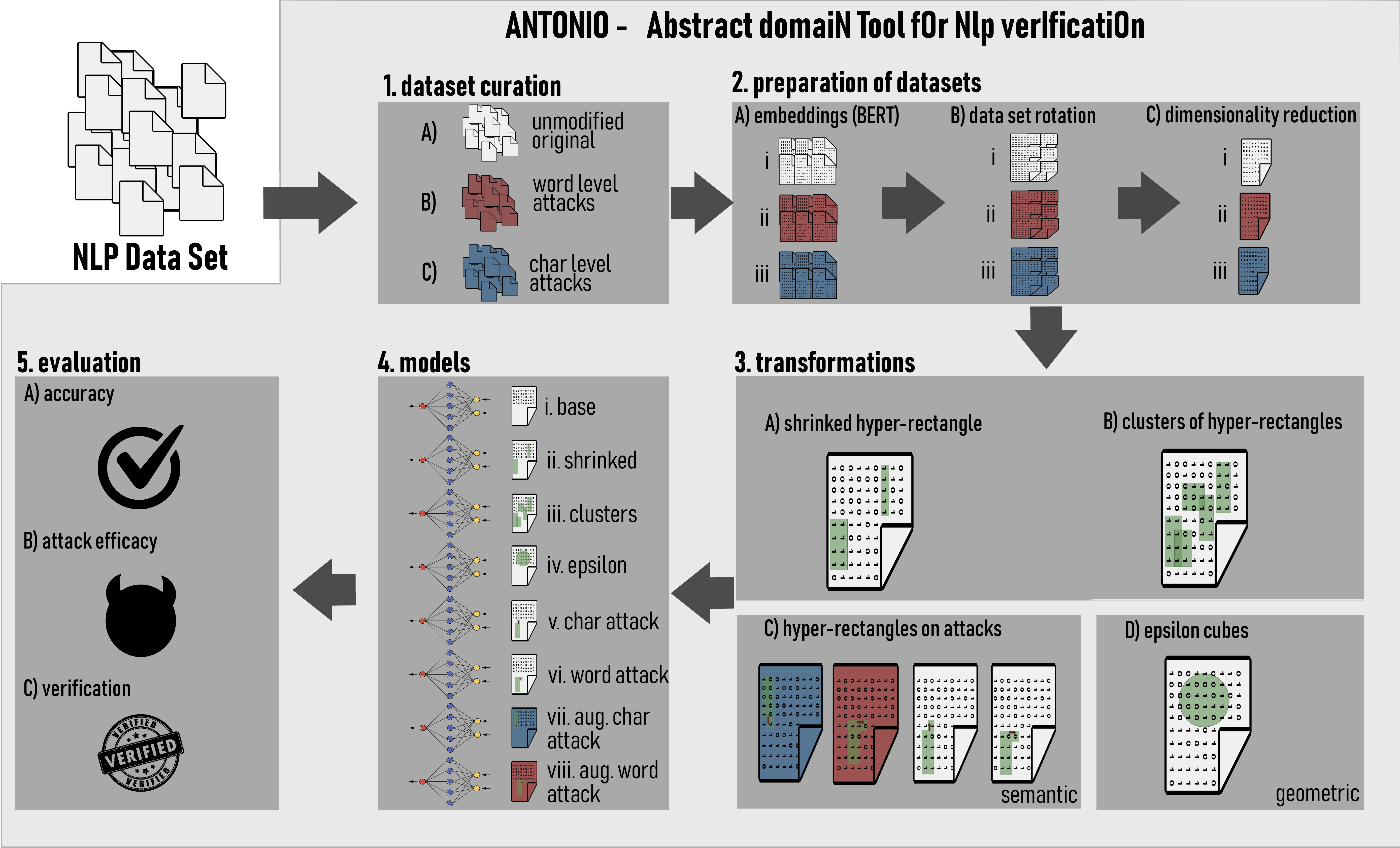}
\caption{\emph{Flow-chart for the tool ANTONIO, showing five main blocks of methods: 1. NLP attacks (``data set curation"); 2.  geometric pre-processing of data, 3. definition of suitable verifiable regions (``transformations"), 4. training models using adversarial training relative to different kinds of attacks, and 5. evaluation of success rates achieved by verifiers and attackers for the trained models.}}
\label{fig:antonio-flow}
\end{figure}

The result is a comprehensive library that contains a toolbox of pre-processing and training methods for verification of NLP models. We call this tool ANTONIO - Abstract domaiN Tool fOr Nlp verIficatiOn (see Figure~\ref{fig:antonio-flow}). 
Although in this tool paper, we evaluate the results on just one R-U-A-Robot dataset, 
the methodology and libraries are completely general, and should work for any NLP dataset and models of comparable sizes. 
We envisage that this work will pave the way for including NLP datasets and problems as benchmarks into DNN verification competitions and papers, and more generally we hope that it will make 
NLP problems more accessible to the neural network verification community.  

\section{ANTONIO and Existing Trends in NLP Verification}
\label{sec:section2}

This paper expands upon the area of NLP verification, this is a nascent field which has only just recently started to be explored.

In their work Zhang et al.~\cite{zhang2021certified} present ARC, a tool that certifies robustness for word level substitution in LSTMs.
They certify robustness by computing an adversarial loss and, if the solution is < 0, then they prove that the model is robust.
However, this tool has limitations such as it certifies only LSTMs that are far below the current state-of-the-art for NLP in terms of size, it uses word embeddings and it is dataset dependant. 
Huang et al.~\cite{huang2019achieving} verify convolutional neural networks using Interval Bound Propagation (IBP) on input perturbations.
However, they only study word synonym and random character replacements on word embeddings.
Ye et al.~\cite{ye2020safer} focus on randomised smoothing for verification of word synonym substitution.
This approach allows them to be model-agnostic, however they are still limited by only 1 type of perturbation and word/sub-word embeddings.
Lastly, Shi et al.~\cite{shi2020robustness} propose a verification method for transformers.
However, their verification property is $\epsilon$-ball robustness and they only demonstrate their method on transformers with less than 3 layers claiming that larger pre-trained models like BERT are too challenging, thus the method is not usable in real life applications.

We can see that although new approaches are being proposed, we have yet to have a consensus on how to approach the verification, and more importantly scalability is a huge issue.
This dictated our exploration of different verification spaces and our focus on filter models as a useful real world verification opportunity.
Geometric shapes, and especially $\epsilon$-balls and $\epsilon$-cubes, are widely used in verification for computer vision to certify robustness properties. In NLP, the aforementioned verification approaches also make use of intervals and IBP, abstract interpretation~\cite{zhang2021certified,huang2019achieving,ye2020safer} and $\epsilon$-ball robustness on word/sub-word embeddings. However, to the best of our knowledge, there is no previous work on geometric/semantic shapes on sentence embeddings.

Table~\ref{tab:verification-comparison} summarises differences and similarities of the above NLP verification approaches. Despite their diversity, most NLP verification approaches make use of a similar methodological pipeline: 

\begin{itemize}
\item Generate “semantic” perturbations on sentences (e.g. the IBP papers tend to use word synonym substitutions, ANTONIO does word, character and sentence level perturbations). 
\item Embed them into continuous spaces (the cited IBP papers use the word embedder GloVe~\cite{pennington2014glove}, ANTONIO uses more modern Sentence-BERT~\cite{reimers2019sentencebert}).
\item  Use geometric bounds (literally bounds in IBP papers or “hyper-rectangles” of different kinds in ANTONIO) to:
      
       \begin{itemize}
       \item Verify the networks' behaviour within those geometric bounds (IBP papers use IBP algorithms to do it, ANTONIO uses off-the shelf tools Marabou and ERAN) and 
      \item Influence neural network training within those input regions. (Custom algorithm is used in IBP papers, ANTONIO uses PGD on defined input space regions).
      \end{itemize}
\end{itemize}

 In the cited IBP papers, the main purpose is to develop a new tailored IBP based method that works for NLP. In it, input region definition is an internal feature of the new IBP algorithms for verification and training. In contrast, in ANTONIO we deliberately disentangle the issues of algorithm design and the problem of input region analysis. This allows us to use state-of-the-art verifiers and training algorithms in a modular fashion; and opens a way for generating benchmarks uniformly.

\begin{table}[tbp]
\centering
\begin{tabularx}{\linewidth}{p{0.1\textwidth}|p{0.09\textwidth}|p{0.11\textwidth}|p{0.13\textwidth}|p{0.14\textwidth}|p{0.09\textwidth}|p{0.12\textwidth}}
\toprule
\textbf{Method} & \textbf{Datasets} & \textbf{NLP attacks} & \textbf{Embeddings} & \textbf{Models} & \textbf{Training algorithm} & \textbf{Verification algorithm} \\
\midrule\midrule
\textbf{ANTO NIO} & RUA Robot, Medical & General purpose: char, word and sentence perturbations & Sentence: Sentence-BERT & Works with any model handled by the verifier & \textbf{PGD}-based & Works with any SoA verifier (\textbf{Marabou}, ERAN) \\
\midrule
Jia et al. (2019) & IMDB, SNLI & Word substitution & Word: GloVe & LSTM, attention-based, CNN, BoW (2 and 3 layers) & \textbf{IBP}-based & IBP-based \\
\midrule
Huang et al. (2019) & AGNews, STT & Char and word substitution & Word: GloVe & CNN (3 and 4 layers) & \textbf{IBP}-based & IBP-based \\
\midrule
Shi et al. (2020) & YELP, STT & $\epsilon$-ball & Word: not specified & Transformers (max 3 layers) & - & Abstract interpretation-based \\
\midrule
Zhang et al. (2021) & IMDB, STT, STT2 & Word perturbations & Word: not specified & LSTM & \textbf{IBP}-based & IBP-based \\
\bottomrule
\end{tabularx}
\caption{\emph{Summary of the main features of the existing NLP verification approaches. In bold are SoA methods.}}
\label{tab:verification-comparison}
\end{table} 

To contrast to verification literature, there is a wide body of work on improving adversarial robustness of NLP systems~\cite{zhang2020adversarial,9557814,wang2021measure,li2021searching,zhou2021defense,zhu2019freelb,dong2021towards}. The approaches previously mentioned make use of data augmentation and adversarial training techniques.
ANOTNIO in the future could be extended with diverse sets of attacks for its defined input regions. 
\section{ANTONIO's Overall Design}
\label{sec:section3}


We now outline the design of the tool ANTONIO (\href{https://github.com/ANTONIONLP/ANTONIO}{github.com/ANTONIONLP/ANTONIO}).
ANTONIO covers every aspect of the NLP verification pipeline, as shown in Figure~\ref{fig:antonio-flow}.
It is modular, meaning that you can modify or remove any part of the NLP verification pipeline, 
which usually consists of the following steps:
\begin{enumerate}
 \item selecting an NLP data set and embedding sentences into vector spaces; 
 \item generating attacks (word, charcter, sentence attacks) on the given sentences, in order to use them for data augmentation, 
 training, or evaluation;
 \item standard machine learning curation for data (e.g. dimensionality reduction) and networks (training, regularisation);
 \item verification, that usually comes with tailored methods of defining input and output regions for networks.  
\end{enumerate}

\noindent ANTONIO is designed to provide support at all of these stages (detailed below):
\paragraph{\textbf{Dataset}} Here we experiment with the R-U-A-Robot dataset, however the user can pick any NLP dataset that can be used for classification. 
 
 \paragraph{\textbf{Semantic attacks and dataset curation}} ANTONIO can create additional augmented datasets by perturbing the original sentences. By incorporating state-of-the-art attacks by~\cite{moradi2021evaluating,wu2021polyjuice}, ANTONIO implements several character-level, word-level, and more sophisticated sentence-level perturbations that can be mixed and matched to create the augmented datasets. Examples of character and word level perturbations  can be found in Tables~\ref{tab:char-perturbations} and~\ref{tab:word-perturbations}.

  \begin{figure}[t]
\centering
	\begin{subfigure}[b]{0.4\textwidth}
		\includegraphics[width=\textwidth]{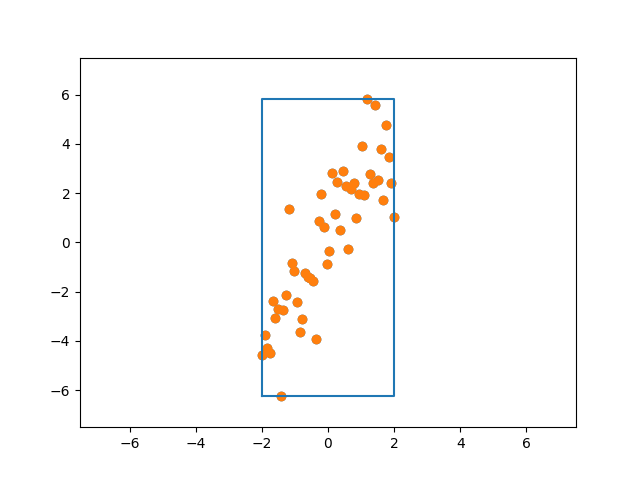}
	\end{subfigure}
	\begin{subfigure}[b]{0.4\textwidth}
		\includegraphics[width=\textwidth]{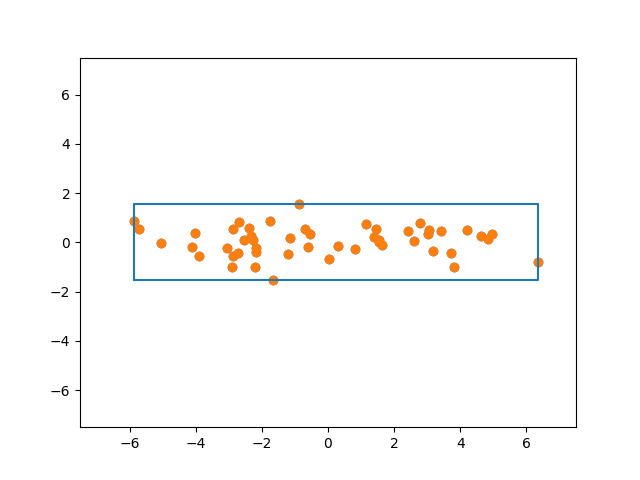}
	\end{subfigure}
\caption{\emph{A 2-dimensional representation of the original data (left) and its eigenspace rotation (right).}}
\label{fig:rotation}
\end{figure}

 \begin{table}[htbp]
	\centering
	\begin{tabularx}{\linewidth}{p{0.15\textwidth}|p{0.35\textwidth}|X|X}
		\toprule
		\textbf{Method}  &  \textbf{Description}  &  \textbf{Original sentence}  &  \textbf{Altered sentence}\\ \midrule\midrule
		Insertion & A character is randomly selected and inserted in a random position. & \emph{Are you a robot?} & \emph{Are yo\textcolor{red}{v}u a robot?}\\ \midrule
		Deletion & A character is randomly selected and deleted. & \emph{Are you a robot?} & \emph{Are you a robt?}\\ \midrule
		Replacement & A character is randomly selected and replaced by an adjacent character on the keyboard. & \emph{Are you a robot?} & \emph{Are you a ro\textcolor{red}{n}ot?}\\ \midrule
		Swapping & A character is randomly selected and swapped with the adjacent right or left character in the word. & \emph{Are you a robot?} & \emph{Are you a r\textcolor{red}{bo}ot?}\\ \midrule
		Repetition & A character in a random position is selected and duplicated. & \emph{Are you a robot?} & \emph{Ar\textcolor{red}{r}e you a robot?}\\ \bottomrule
	\end{tabularx}
	\caption{\emph{Character-level perturbations: their types and examples of how each type acts on a given sentence from the R-U-A-Robot dataset~\cite{gros2021ruarobot}.  Perturbations are selected from random words that have 3 or more characters, first and last characters of a word are never perturbed.}}
	\label{tab:char-perturbations}
	
\end{table}

 \begin{table}[htbp]
	\centering
	\begin{tabularx}{\linewidth}{p{0.15\textwidth}|p{0.35\textwidth}|X|X}
		\toprule
		\textbf{Method} & \textbf{Description} & \textbf{Original sentence} & \textbf{Altered sentence} \\ \midrule \midrule
		Deletion & Randomly selects a word  and removes it. & \footnotesize{\emph{Can u tell me if you are a chatbot?}} & \emph{Can u tell if you are a chatbot?}\\ \midrule
		Repetition & Randomly selects a word and duplicates it. & \emph{Can u tell me if you are a chatbot?} & \emph{Can \textcolor{red}{can} u tell me if you are a chatbot?}\\ \midrule
		Negation &  Identifies verbs then flips them (negative/positive). & \emph{Can u tell me if you are a chatbot?} & \emph{Can u tell me if you are \textcolor{red}{not} a chatbot?}\\ \midrule
		Singular/ plural verbs & Changes verbs to singular form, and conversely. & \emph{Can u tell me if you are a chatbot?} & \emph{Can u tell me if you \textcolor{red}{is} a chatbot?}\\ \midrule
		Word order &  Randomly selects consecutive words  and changes the order in which they appear.  & \emph{Can u tell me if you are a chatbot?} & \emph{Can u tell me if you are \textcolor{red}{chatbot a}?}\\ \midrule
		Verb tense & Converts present simple or continuous verbs to their corresponding past simple or continuous form. & \emph{Can u tell me if you are a chatbot?} & \emph{Can u tell me if you \textcolor{red}{were} a chatbot?}\\ \bottomrule
	\end{tabularx}
	\caption{\emph{Word-level perturbations: their types and examples of how each type acts on a given sentence from the R-U-A-Robot dataset~\cite{gros2021ruarobot} .}}
	\label{tab:word-perturbations}
\end{table}

\paragraph{\textbf{Dataset preparation}} This block can be considered as geometric data manipulations. First of all, we need an embedding function. ANTONIO utilises SentenceBERT~\cite{reimers2019sentencebert} as a sentence embedder. The model  implemented within ANTONIO produces embeddings in 384 dimensions. The user can, however, substitute SentenceBERT with any embedding function they prefer.
 The original part of ANTONIO implements data rotation to help the hyper-rectangles to better fit the data (as shown in Figure~\ref{fig:rotation}).
 Lastly, we use PCA for dimensionality reduction~\cite{scikit-learn}. This helps verification algorithms to reduce over-approximation and speeds up training and verification by reducing the input space.
 These last two data manipulations can be arbitrarily omitted or modified and the user can insert other manipulations of their choices that might help.

\begin{figure}[t]
\centering
	\begin{subfigure}[b]{0.3\textwidth}
		\includegraphics[width=\textwidth]{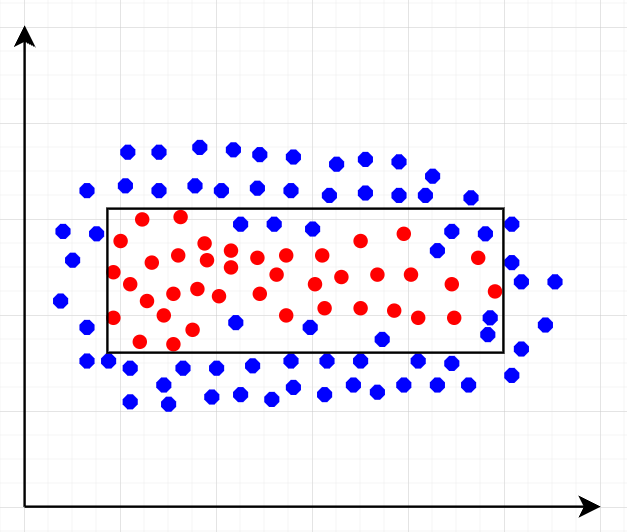}
	\end{subfigure}
	\begin{subfigure}[b]{0.3\textwidth}
		\includegraphics[width=\textwidth]{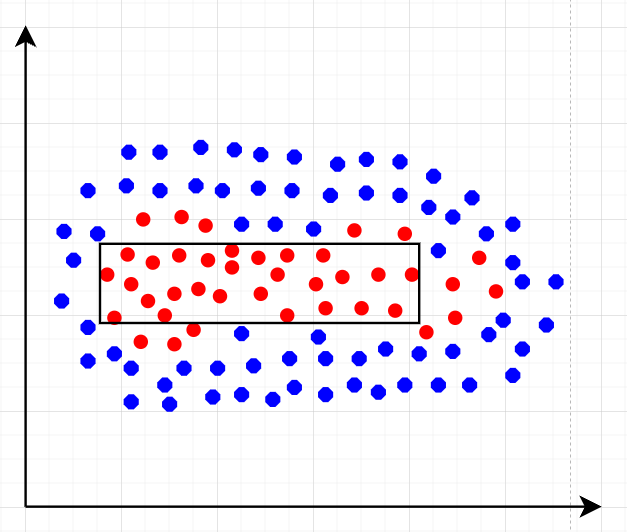}
	\end{subfigure}
	\begin{subfigure}[b]{0.3\textwidth}
		\includegraphics[width=\textwidth]{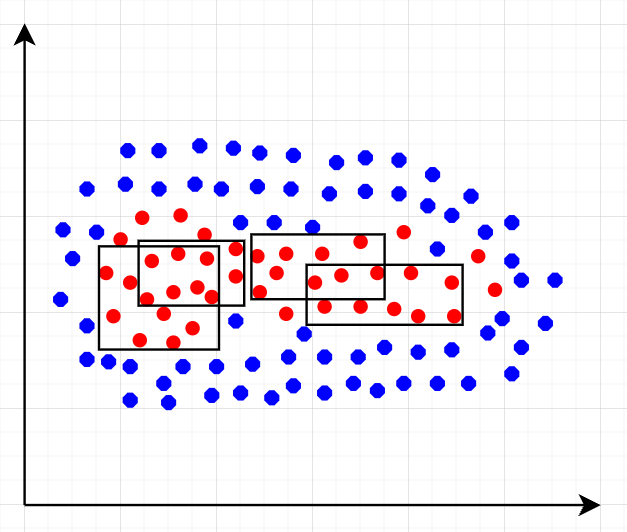}
	\end{subfigure}
\caption{\emph{An example of hyper-rectangle (left), shrunk hyper-rectangle (centre) and clustered hyper-rectangles (right) in 2-dimensions. The red dots represent sentences in the embedding space of one class, while the blue dots are embedded sentences that do not belong to that class.}}
\label{fig:shrink-cluster}
\end{figure}

\paragraph{\textbf{Hyper-rectangles}} The original core of ANTONIO as a tool is the code producing hyper-rectangles for given sentences (and their semantic attacks). We implemented several ways to create and refine the hyper-rectangles to increase their precision. 
 Figure~\ref{fig:shrink-cluster} (left) shows how a naive hyper-rectangle, that contains all the inputs from the desired class, might also contain inputs from the other class. That is why we implemented a method to shrink the hyper-rectangle to exclude the undesired inputs (centre) and a method for clustering and generating multiple hyper-rectangles around each cluster (right).
 Furthermore, to increase precision, ANTONIO can create a third type of hyper-rectangles (Figure~\ref{fig:antonio-flow} (3C)) by attacking the inputs (as described in the data augmentation part) and then drawing the hyper-rectangles around each input and its perturbations.
 Finally, for comparison purposes, we also implemented the creation of $\epsilon$-cubes.
 These hyper-rectangles will be used both for training and verification.

\paragraph{\textbf{Training}} We implemented three methods for training: base training, data augmentation, and adversarial training
 The base models are trained with a standard cross-entropy loss on the original dataset.
 The data augmentation models are still trained with a standard cross-entropy loss but on the augmented datasets.
 For adversarial training, instead, we are using Projected Gradient Descent (PGD)~\cite{madry2019deep} to calculate the worst case perturbation for each input on each epoch and we are adding those perturbations when calculating the loss.
 Usually, PGD perturbations are projected back into the given $\epsilon$-cube.  ANTONIO projects within the given hyper-rectangles instead.
 The user can choose any combination of training methods, hyper-rectangles, and attacks to train the models or they can implement new ones as well. Section~\ref{sec:section4} will illustrate this modular usage of the tool.
 
\paragraph{\textbf{Evaluation and Verification}} For the evaluation, ANTONIO implements mainly three metrics.
 Firstly it simply calculates the standard accuracy of the model, as it is important to not have a significant drop in accuracy when you train a model for robustness.
 Secondly, it computes robustness to attack (accuracy on adversarial test samples), which is obtained by generating several perturbations of the test set and by calculating accuracy on those.
 Finally, ANTIONIO calculates the percentage of verified hyper-rectangles. For testting purposes we  connected ANTONIO to two state-of-the-art verifiers: ERAN~\cite{singh2019abstract,adcnn} and Marabou~\cite{marabou}. 
 Thanks to VNN-COMP, the annual neural network verification competition~\cite{muller2022third}, the community developed common neural network verification standards, that require ONNX format for neural networks~\cite{bai2019} and VNNLIB format for verification queries~\cite{vnnlib2022}. ANTONIO automatically performs translation of neural networks into ONNX, and specifications of hyper-rectangles in Python -- into the  VNNLIB format.      
 Furthermore, ANTONIO implements methods for generating queries, retrieving the data and calculating the statistics on them.
 The user can connect and use any verification tool that they prefer and also add any other metric of choice.



\section{Example of Benchmarking with ANTONIO}
\label{sec:section4}

\begin{table}[htbp]
\centering
\begin{tabular}{l|l|rrr}
\toprule
\textbf{Verifier} & \textbf{Model} & \textbf{$\mathbb{H}^*_{\epsilon=0.05}$} & \textbf{$\mathbb{H}^*_{char}$} & \textbf{$\mathbb{H}^*_{word}$}\\
\midrule\midrule
\multirow[c]{4}{*}{Marabou}
& $N_{base}$ & 1.79 & 4.88 & 11.69 \\
& $N_{\epsilon=0.05}$ & \textbf{18.46} & 21.99 & 41.93 \\
& $N_{char-adv}$ & 7.37 & \textbf{30.41} & 41.93 \\
& $N_{word-adv}$ & 12.17 & 25.82& \textbf{45.12} \\
\midrule
\multirow[c]{4}{*}{ERAN}
& $N_{base}$ & 0.00 & 0.87 & 1.80 \\
& $N_{\epsilon=0.05}$ & 0.12 & 4.18 & 10.16 \\
& $N_{char-adv}$ & 0.00 & 4.43 & 8.97 \\
& $N_{word-adv}$ & 0.04 & 4.05 & 10.75 \\
\bottomrule
\end{tabular}
\caption{\emph{Example how ANTONIO-generated networks and input shapes form a uniform benchmark for two verifiers.}}
\label{tab:verification-comparison}
\end{table}

We now show how to use ANTONIO for verification benchmarking, using the R-U-A-Robot dataset as an example.

Following the pipeline described in Section~\ref{sec:section3}, ANTONIO generates a set of character, word and sentence perturbations for the dataset, by perturbing each sentence several times in turn. 
 Tables~\ref{tab:char-perturbations} and~\ref{tab:word-perturbations} show how, given one sentence and one type of perturbation, one forms a set of perturbed sentences.
ANTONIO keeps different types of perturbations separately.  We obtain, per each sentence, three sets of its perturbed variants, one per each perturbation type.

ANTONIO embeds these sentences into a vector space, and performs 
the geometric transformations, such as data rotation and dimensionality reduction. 
Then, for each perturbation type, ANTONIO creates one hyper-rectangle per each original sentence. The hyper-rectangle covers the original sentence and its $n$ perturbations. ANTONIO collects all such hyper-rectangles for the given set of sentences. We call the resulting sets of hyper-rectangles $\mathbb{H}^*_{char}$ and $\mathbb{H}^*_{word}$, $\mathbb{H}^*_{sentence}$, depending on the perturbation type. Similarly, it also creates a set of $\epsilon$-cubes (denoted $\mathbb{H}^*_{\epsilon=0.05}$), based on the $L_{\infty}$ distance, and ignoring any knowledge of semantic perturbations.
Finally, ANTONIO trains a baseline network ($N_{base}$) and four adversarial networks ($N_{\epsilon=0.05}$, $N_{word-adv}$, $N_{char-adv}$, $N_{sentence-adv}$ ) trained using PGD on, respectively, $\mathbb{H}^*_{\epsilon=0.05}$,  $\mathbb{H}^*_{word}$, $\mathbb{H}^*_{char}$, $\mathbb{H}^*_{sentence}$.
Having made these preparations, ANTONIO forms a set of verification challenges, to verify the given networks and the input space covered by $\mathbb{H}^*_{\epsilon=0.05}$, $\mathbb{H}^*_{char}$, $\mathbb{H}^*_{word}$ and $\mathbb{H}^*_{sentence}$.
At this point, the automatically generated neural networks in ONNX format and the verification conditions defining the desirable shapes in VNNLib format can be given to any verifier participating in VNN-COMP. 
Table~\ref{tab:verification-comparison} shows the results for Marabou and ERAN, the 2 verifiers that we tested with ANTONIO already. But this list can be easily extended. For brevity, we note the results presented (Table~\ref{tab:verification-comparison}) are just a snapshot of what ANTONIO is capable of, and we omit experiments done on sentence perturbations.



\section*{Acknowledgements}
Authors acknowledge support of EPSRC grant AISEC EP/T026952/1 and NCSC grant Neural Network Verification: in search of the missing spec.


\bibliographystyle{plain}
\bibliography{references}

\clearpage

\end{document}